\documentclass[lettersize,onecolumn]{IEEEtran}
\usepackage{amssymb}
\usepackage{amsmath}
\usepackage{algorithm}       
\usepackage{algpseudocode}
\usepackage{subcaption}
\usepackage{amsmath}
\usepackage{amsmath}
\usepackage{amsfonts}
\usepackage{amssymb}
\usepackage{newfloat}
\usepackage{listings}
\usepackage{amsmath}
\usepackage{booktabs} 
\usepackage{xcolor}
\usepackage{colortbl}
\usepackage{amssymb}
\usepackage{booktabs} 
\usepackage{hyperref}
\usepackage{graphicx}     
\usepackage{xcolor}       
\usepackage{array} 
\usepackage{algorithm}
\usepackage{amsmath}
\usepackage{amsfonts}
\usepackage{amssymb}
\usepackage{amsthm}
\usepackage{amsmath}
\usepackage{amsmath}
\usepackage{bm}
\usepackage{bibentry}
\usepackage{lineno}
\floatstyle{ruled}

\begin{document}

\title{A Lightweight 3D Anomaly Detection Method with Rotationally Invariant Features}

\author{Hanzhe~Liang, \thanks{Hanzhe Liang is with the College of Computer Science and Software Engineering, Shenzhen University, Shenzhen 518060, China; also with the Shenzhen Audencia Financial Technology Institute, Shenzhen University, Shenzhen 518060, China; and also with Audencia Nantes École de Management, Nantes 44300, France (e-mail: lianghanzhe2023@email.szu.edu.cn).}%
Jie~Zhou, \thanks{Jie Zhou is with the School of Mathematics and Statistics, Changsha University of Science and Technology, Changsha 410114, China.}%
Can~Gao$^{\dagger}$, \thanks{Can Gao is with the College of Computer Science and Software Engineering, Shenzhen University, Shenzhen 518060, China; and also with the Guangdong Provincial Key Laboratory of Intelligent Information Processing, Shenzhen 518060, China (e-mail: davidgao@szu.edu.cn).}%
Bingyang~Guo, \thanks{Bingyang Guo is with the Software College, Northeastern University, Shenyang 110819, China.}%
Jinbao~Wang,
and~Linlin~Shen \thanks{Jinbao Wang and Linlin Shen are with the School of Artificial Intelligence, Shenzhen University, Shenzhen 518060, China (e-mail: wangjb@szu.edu.cn; llshen@szu.edu.cn).}
}

\maketitle

\begin{abstract}
3D anomaly detection (AD) is a crucial task in computer vision, aiming to identify anomalous points or regions from point cloud data. However, existing methods may encounter challenges when handling point clouds with changes in orientation and position because the resulting features may vary significantly. To address this problem, we propose a novel Rotationally Invariant Features (RIF) framework for 3D AD. Firstly, to remove the adverse effect of variations on point cloud data, we develop a Point Coordinate Mapping (PCM) technique, which maps each point into a rotationally invariant space to maintain consistency of representation. Then, to learn robust and discriminative features, we design a lightweight Convolutional Transform Feature Network (CTF-Net) to extract rotationally invariant features for the memory bank. To improve the ability of the feature extractor, we introduce the idea of transfer learning to pre-train the feature extractor with 3D data augmentation. Experimental results show that the proposed method achieves the advanced performance on the Anomaly-ShapeNet dataset, with an average P-AUROC improvement of 17.7\%, and also gains the best performance on the Real3D-AD dataset, with an average P-AUROC improvement of 1.6\%. The strong generalization ability of RIF has been verified by combining it with traditional feature extraction methods on anomaly detection tasks, demonstrating great potential for industrial applications.
\end{abstract}



\section{Introduction}

Three Dimension anomaly detection (3D-AD)~\cite{zhu2024towards, li2023scalable3danomalydetection} has attracted increasing attention from the computer vision community due to its widespread application in inspecting high-precision industrial products. It is dedicated to identifying anomalous points or regions that deviate from given 3D point cloud data. In real-world industrial applications, we often face a scenario where it is easy to collect normal samples but time-consuming and costly to acquire defective samples, such that unsupervised methods that only learn from normal samples are widely used.

For 3D AD, some traditional methods such as BTF~\cite{horwitz2022featureclassical3dfeatures,10898004} have been proposed to extract features directly from 3D structures, but they have limitations in extracting cross-sample features to accommodate the diversity of different samples. In view of the great success of deep learning, deep neural network-based methods have become the mainstream approach for 3D AD. These methods can be roughly categorized into feature embedding-based~\cite{Liu2023real3d} and reconstruction-based~\cite{zhou2025r3dad}. 

Feature embedding-based methods use pre-trained models to extract features of normal samples for a memory bank, with anomalies detected by comparing test sample features to those in the bank. Some methods such as M3DM~\cite{Wang2023multimodal} have achieved on the datasets using memory bank methods. However, they face challenges on unregistered datasets such as Real3D-AD~\cite{Liu2023real3d} and Anomaly-ShapeNet~\cite{li2023scalable3danomalydetection}. Reg3D-AD~\cite{Liu2023real3d} used PointMAE~\cite{Pang2022pointmae} for feature representation based on the registered point clouds, and Group3AD~\cite{zhu2024towards} further used contrast learning to improve feature differentiation ability.
On the other hand, reconstruction-based methods encode point cloud data into a latent space and decode it in its original form, with high reconstruction errors indicating anomalies. IMRNet~\cite{li2023scalable3danomalydetection} improved anomaly detection by using geometric-preserving downsampling and random masking. R3D-AD~\cite{zhou2025r3dad} utilized PointNet~\cite{Qi_2017_CVPR} and a diffusion process to progressively restore the point cloud. These methods have shown promising results on the Anomaly-ShapeNet~\cite{li2023scalable3danomalydetection} and Real3D-AD~\cite{Liu2023real3d} datasets.

\begin{figure}[!th]
    \centering
    \includegraphics[width=0.6\linewidth]{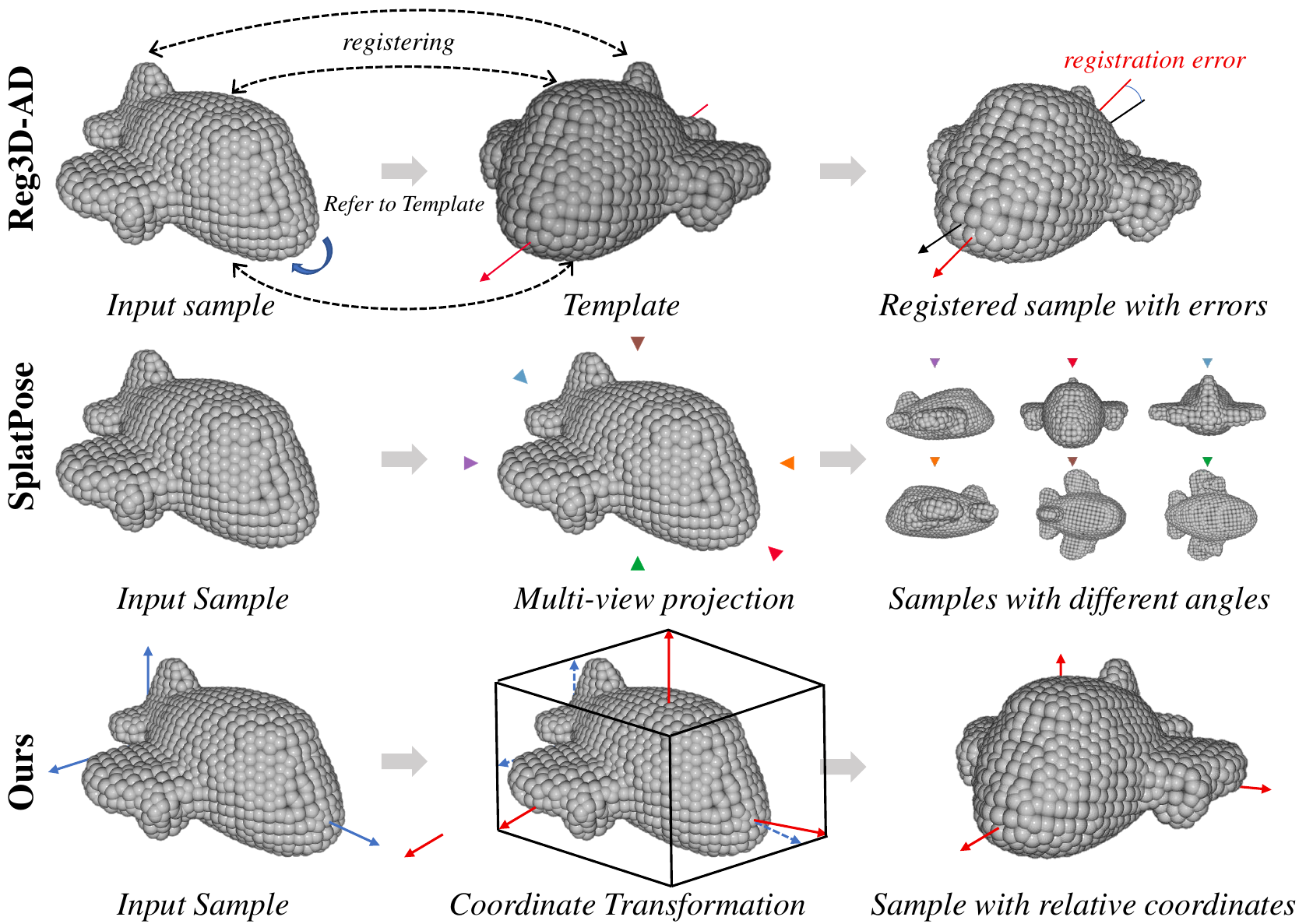}
    \caption{Comparison of representative methods for point clouds with variations in orientation and position. Reg3D-AD~\cite{Liu2023real3d} aligns the testing sample with the standard template by registration. SplatPose~\cite{Kruse2024spkatpose} reconstructs 3D object from multi-view images by Gaussian splash. Our method maps point clouds into a rotational invariance space by coordinate transformation.}
    \label{motivation}
\end{figure}

Industrial 3D AD is particularly difficult because products on the production line may appear in different poses~\cite{cheng2025highresolution3danomalydetection}. Consequently, the same product may have point cloud data with variations in orientation and position. Although a rotating point cloud does not alter its geometric structure, it changes the coordinate representation corresponding to the structure, resulting in a significant difference in the output of the feature extraction network. Reg3D-AD~\cite{Liu2023real3d} aligns test samples with a standard template via registration to reduce data variation effects, but this method is computationally intensive and risks registration errors, particularly with significant train-test gaps. 
{SplatPose~\cite{Kruse2024spkatpose} leverages novel view synthesis via Gaussian splatting to reconstruct 3D objects and enhance pose robustness; however, its point cloud fidelity still depends heavily on image quality, synthesized-view consistency, and view count, and it requires test-time image–3D alignment. These limitations motivate us to pursue a purely geometric and view-free representation that preserves rotation invariance directly on point clouds.}
Therefore, our method proposed to map point clouds into a rotational invariance space through coordinate transformation, enabling the generation of a unique feature representation for point clouds with different rotations and translations. The differences between our solution and previous methods are shown in Figure~\ref{motivation}. 
\label{sec:intro}

Motivated by the above facts, we present the Rotationally Invariant Features~(RIF) framework. Specifically, we first introduce a Point Coordinate Mapping (PCM) technique to transform point cloud coordinates into the relative coordinate system.
Then, we design a Convolutional Transform Feature Network~(CTF-Net) based on Compositional Convolution, capturing local point cloud features at different scales. To assist the pre-training of CTF-Net, we develop a Spatial 3D Data Augmentation~(S3DA) technique to diversify training samples for better representation. Through comparative experiments and ablation studies on Anomaly-ShapeNet and Real3D-AD, it is observed that our framework is highly appealing and achieves state-of-the-art results in various metrics. To the best of our knowledge, we are the first method to employ a deep feature extractor for the rotationally invariant feature representation of point clouds. The contributions are summarized as follows:

\begin{itemize}
    \item To eliminate the adverse effects of point cloud data variations on feature extraction, we develop a point coordinate mapping~(PCM) technique, which maps point clouds to a rotation-invariant space, enabling consistent point coordinate representation for downstream feature extraction.
    
    
    \item To effectively extract point cloud features, we design a Convolutional Transform Feature Network~(CTF-Net), which performs multiple 1D convolutions at different scales to capture local structure information. Additionally, we introduce the Spatial 3D Data Augmentation~(S3DA) strategy to aid the pre-training of our feature extraction network, providing diverse training data to enhance feature representation.
    
    \item Based on PCM and CTF-Net, we propose the Rotationally Invariant Features~(RIF) framework for anomaly detection. Our framework can seamlessly integrate with other point cloud feature extractors, providing a general and practical solution for 3D anomaly detection.
    
    
    \item We conduct extensive comparative and ablation experiments on the Anomaly-ShapeNet and Real3D-AD datasets. Experimental results show the superiority of our method, with a P-AUROC improvement of 17.7\% on Anomaly-ShapeNet and 1.6\% on Real3D-AD.
\end{itemize}

The rest of this paper is organized as follows. Section~\ref{relatedwork} recalls some related work. Section~\ref{Approch} details our proposed method. In Section~\ref{experiment}, extensive experiments are conducted to validate the proposed method. Finally, Section~\ref{Conclusion} draws the conclusion of the paper.

\section{Related Work}
\label{relatedwork}
\subsection{ 2D Anomaly Detection}
2D anomaly detection~\cite{ Zhao2023omnial,LEE2026111820,LIU2026112198,LI2024110258,MIAO2026112163,tang2025mpr} is an important vision task that aims to detect and localize anomalies from images. Existing 2D AD methods are mainly classified into two categories: generative and discriminative methods~\cite{PangSC2021,10034849}. Generative models such as Autocoder~\cite{Gong2019memae}, GAN, and Diffusion are employed to learn the feature representation or distribution of normal samples, and anomalies are detected by performing image reconstruction~\cite{ZAVRTANIK2021reconstruction,luo2025exploring} or matching with a memory bank~\cite{roth2022totalrecallindustrialanomaly,luo2025ura}. Recent methods, such as DRAEM~\cite{Zavrtanik2021DRAEM}, PatchCore~\cite{roth2022totalrecallindustrialanomaly} and CRAD have yielded impressive results. Discriminative methods aim to train a supervised model to discriminate anomalous and normal images, which usually requires a certain number of anomalous samples annotated by humans or generated by data augmentation. Some methods BGAD~\cite{Zhang2023prototypical}, and AHL~\cite{Zhu2024openset} have shown the benefits of supervised information to improve 2D anomaly detection performance. Although promising results, applying 2D  anomaly detection methods to 3D point clouds directly may face significant challenges.


\subsection{3D Anomaly Detection}
3D anomaly detection intends to identify and locate anomalous points or regions within 3D point cloud data~\cite{3Dsurvey,liang2025taming}. Existing methods for 3D AD can be grouped into two main categories: feature embedding-based methods and reconstruction-based methods. Feature embedding-based methods use pre-trained models to extract normal features for a memory bank, detecting anomalies by comparing test features against stored entries~\cite{10702559,ZHANG2025111360,1asda,10898004}. Reg3D-AD~\cite{Liu2023real3d,10260338} uses PointMAE~\cite{Pang2022pointmae} to extract global geometric and local coordinate features post-registration, matching both during inference for anomaly scoring. {Group3AD~\cite{zhu2024towards} clusters groups into uniformly compact structures, constructing group-level features for the memory bank. Reconstruction-based methods aim to project point cloud data into a high-dimensional latent space through encoding processes, subsequently reconstructing the original input via decoding, with points demonstrating elevated reconstruction errors being identified as anomalies. 
PASDF~\cite{zheng2025bridging} learns high-quality continuous representations of the signed distance for each point and directly identifies outlier anomalies based on the signed distances.}
IMRNet~\cite{li2023scalable3danomalydetection} improved PointMAE by geometric-preserving downsampling and random masking to enhance reconstruction ability for anomaly detection.
R3DAD~\cite{zhou2025r3dad} employed PointNet to progressively restore full-mask point clouds by performing the diffusion, aiding the model in accurately identifying abnormal regions. Additionally, some methods integrate 2D images with 3D point clouds to perform RGB-D AD. M3DM~\cite{Wang2023multimodal} aligns PointTransformer (3D) and ViT (2D) features via contrastive learning.
CPMF~\cite{cao2023complementarypseudomultimodalfeature} combines local 3D geometry with 2D semantics from multi-view images.
Looking3D~\cite{Bhunia2024look} integrates texture-enhanced 3D data with cross-modality matching.

In 3D point cloud AD, data variations in orientation significantly affect detection performance. IMRNet~\cite{li2023scalable3danomalydetection} and R3DAD \cite{zhou2025r3dad} used data augmentation techniques to introduce diversity through performing geometric transformations, noise injection, and random masking on point clouds. Reg3D-AD~\cite{Liu2023real3d} and Group3AD~\cite{zhu2024towards} introduced registration to align different point cloud samples to improve feature extraction. CPMF~\cite{cao2023complementarypseudomultimodalfeature}, Looking3D~\cite{Bhunia2024look}, and SplatPose~\cite{Kruse2024spkatpose} employed multi-view strategies to capture features from different angles to enhance the robustness of anomaly detection. Although these methods partially alleviate the adverse effect of orientation on detection performance, there is still a strong desire to develop a simple and computationally efficient technique to accommodate 3D point cloud data with arbitrary orientation.

{Traditional descriptors such as FPFH and multi-FPFH rely on statistical histograms of local geometric relations, providing only implicit rotation robustness. In contrast, PCM constructs an explicit rotation-invariant coordinate frame via deterministic geometric anchoring and orthogonalization, making its mathematical formulation fundamentally different from these histogram-based operators.}

\section{Approach}
\label{Approch}

\begin{figure*}[!ht]
    \centering
\includegraphics[width=1\linewidth]{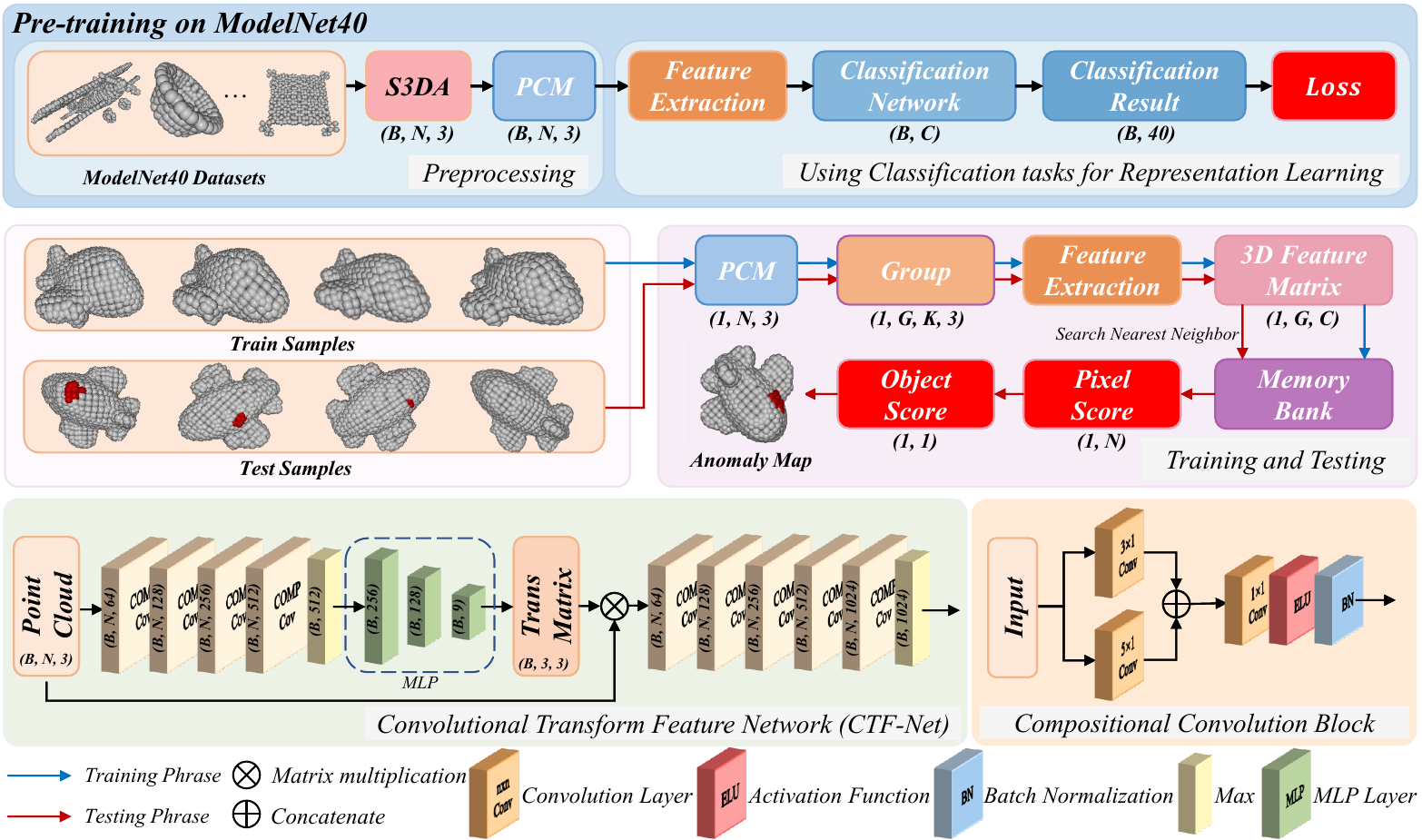}
    \caption{Pipeline of our RIF framework. Feature extractor CTF-Net is first pre-trained on the ModelNet40 dataset with 3D data augmentation and coordinate mapping. Then, normal training samples from the target dataset are mapped and grouped for the pre-trained CTF-Net to extract discriminative features for a memory bank. Finally, the extracted features of testing samples are matched with the memory bank to generate pixel and object scores by nearest neighbor searching.}
    \label{pipline}
\end{figure*}

The core of our RIF framework is to learn rotationally invariant features to address the diversity of samples in orientation, position, and posture. Here, we propose a simple yet efficient solution for 3D AD, which is shown in Figure~\ref{pipline}. We will elaborate on its details in the following sections. 
To improve the understanding, the main symbols in the paper and their explanations are reported in Table~\ref{sy}.

\begin{table*}[!ht]         
\centering
\small
\begin{tabular}{p{3.2cm}p{10cm}}
\hline
\textbf{Notation} & \textbf{Description} \\ 
\hline
$\mathbf{P}_{(n,3)}$ & Raw point cloud containing $n$ 3D points \\
$\mathcal{F}: \mathbf{P}_{(n,3)} \rightarrow \mathbf{F}_{(n',d)}$ & Feature extractor mapping a point cloud to $d$-dimensional features \\
$\mathbf{c} \in \mathbb{R}^3$ & The centroid of a given point cloud: $\mathbf{c} = \frac{1}{n}\sum_{i=1}^n \mathbf{p}_i$ \\
$\mathbf{u}_1,\mathbf{u}_2,\mathbf{u}_3 \in \mathbb{R}^3$ & Basis vectors: $\mathbf{u}_1 = \mathbf{p}_{\text{far1}}-\mathbf{c}$, $\mathbf{u}_2 = \mathbf{p}_{\text{far2}}-\mathbf{c}$, $\mathbf{u}_3 = \mathbf{p}_{\text{near}}-\mathbf{c}$ \\
$\mathbf{e}_1,\mathbf{e}_2,\mathbf{e}_3 \in \mathbb{R}^3$ & Orthonormal basis via Gram-Schmidt orthogonalization \\
$\mathbf{p}_i^* \in \mathbb{R}^3$ & Mapped point coordinates: $\mathbf{p}_i^* = (\mathbf{p}_i - \mathbf{c})[\mathbf{e}_1;\mathbf{e}_2;\mathbf{e}_3]^T$ \\
$S(\cdot), J(\cdot), Z(\cdot)$ & S3DA operations: anisotropic scaling, jittering, zero-masking \\
$\mathbf{V}_{(1,512)} \in \mathbb{R}^{512}$ & Global feature vector from CTF-Net \\
$\mathbf{M}_{3\times3} \in \mathbb{R}^{3\times3}$ & Transformation matrix generated by MLP \\
$\mathbf{F}^*_{(1,1024)} \in \mathbb{R}^{1024}$ & Final sample-level feature vector \\
$G \in \mathbb{N}^+$ & Number of groups (default: 512) \\
$K \in \mathbb{N}^+$ & Number of points per group (default: 512) \\
$\mathbf{Mem}$ & Memory bank storing normal features \\
$\|\cdot\|_2$ & $L_2$-norm for nearest neighbor distance \\
P-AUROC & Pixel-level Area Under ROC Curve \\
O-AUROC & Object-level Area Under ROC Curve \\
P-AUPRO & Pixel-level Area Under PRO Curve \\
O-AUPRO & Object-level Area Under PRO Curve \\
\hline
\end{tabular}
\caption{Main symbols and their explanations in the paper.}
\label{sy}
\end{table*}

\subsection{Point Coordinate Mapping for Rotational Invariance}
\label{PCM}
For 3D point cloud data, maintaining rotational invariance is crucial for ensuring model robustness. Existing feature extractors try to learn representative features from each sample, i.e., $\mathcal{F}: \textbf{P}_{(n, 3)} \rightarrow \textbf{F}_{(n^{\prime}, d)}$, where $\textbf{P}_{(n, 3)}$ denotes the set of 3D points with the number of $n$, and $\textbf{F}_{(n^{\prime}, d)}$ represents the learned feature map with the size of $n^{\prime} * d$. However, in this process, any rotation imposed on the input point cloud may alter the output features, affecting the robustness of features and performance of the trained model. To address this problem, we propose a PCM module to transform point cloud samples. Specifically, the PCM aims to construct a relative coordinate representation for each sample with any rotation and translation. It calculates the centroid \(\mathbf{c}\) of each sample: 
$\mathbf{c} = \frac{1}{n} \sum_{i=1}^n \mathbf{p}_i$,
where $\mathbf{p}_i$ is a point within the sample, and $n$ is the number of all points. Then, PCM sequentially selects three key vectors to define the relative coordinate system in the following manner: 

\begin{itemize}
    \item \(\mathbf{u}_1 = \mathbf{p}_{\text{far1}} - \mathbf{c}\), where \(\mathbf{p}_{\text{far1}}\) is the farthest point from the centroid;
    \item \(\mathbf{u}_2 = \mathbf{p}_{\text{far2}} - \mathbf{c}\), where \(\mathbf{p}_{\text{far2}}\) is the second farthest point from the centroid, and is linearly independent to \(\mathbf{u}_1\);
    \item \(\mathbf{u}_3 = \mathbf{p}_{\text{near}} - \mathbf{c}\), where \(\mathbf{p}_{\text{near}}\) is the nearest point to the centroid that makes the rank of the three selected vectors equal to 3, i.e., $Rank([\mathbf{u}_1;$ $\mathbf{u}_2; \mathbf{u}_3]) = 3$;
    {\item If multiple farthest or nearest points satisfy $\|\mathbf{p}-\mathbf{c}\|=d_{\max}$ or $d_{\min}$, we replace them with their centroid to resolve distance ties, yielding a deterministic representative point and a stable PCM basis.}

\end{itemize}

It is well-known that spanning a 3D space requires at least three vectors. In case the selected vectors are linearly dependent, PCM sequentially selects the next farthest or nearest points until the constraints are satisfied. After determining these three vectors, Schmidt orthogonalization is performed to generate standard orthogonal basis vectors  \(\mathbf{e}_1\), \(\mathbf{e}_2\), and \(\mathbf{e}_3\). For each point, it can be transformed into the relative coordinate point by: 
\begin{equation}
\label{Point_Mapping}
\mathbf{p}_i^{*} = (\mathbf{p}_i - \mathbf{c}) [\mathbf{e}_1; \mathbf{e}_2; \mathbf{e}_3]^{\text{T}}.
\end{equation}

In fact, a point cloud sample with any rotation and translation can be mapped into a relative coordinate representation by the PCM: $\mathcal{M}: \textbf{P}_{(n, 3)} \rightarrow \textbf{P}_{(n, 3)}^{*}$ (See \textit{Appendix} for the theoretical proof). This mapping ensures that each sample with different rotations and translations has the same coordinate representation, providing a stable input for downstream feature extraction. The implementation of PCM can be depicted by Algorithm~\ref{PCMp}.

\subsection{Pre-trained Feature Extractor with Spatial 3D Data Augmentation}
Due to the limited number of training samples in 3D anomaly detection, learning a good feature extractor may face great challenges. To enhance the feature extraction ability for 3D point cloud data, we resort to the technique of network pre-training and transferring. Specifically, the feature extractor is pre-trained from scratch on the ModelNet40 dataset.  
To enhance data diversity and feature generalization, we introduce Spatial 3D Data Augmentation~(S3DA) as a complementary approach to PCM. It diversifies training data by applying additional augmentation techniques to introduce spatial variability without compromising consistency, while PCM realizes rotational invariance through mapping points into relative coordinate representation. S3DA achieves this by combining several augmentation strategies, including random scaling along different axes, point-level random jittering, and random zeroing of specific points. These operations can be represented as:
\begin{equation}
\textbf{P}_{(n, 3)}^{\prime} = Z(J(S(\textbf{P}_{(n, 3)}))),
\end{equation}
\noindent where $\textbf{P}_{(n, 3)}$ and $\textbf{P}_{(n, 3)}^{\prime}$ denote the original and augmented pointclouds, respectively, $S(\cdot)$, $J(\cdot)$, and $Z(\cdot)$ mean the random scaling, jittering, and zeroing, respectively.

After augmentation and mapping, the preprocessed ModelNet40~\cite{sun2022modelnet40} dataset is used to pre-train the feature extractor to learn the generalized representation ability on the classification task. By pre-training on the preprocessed 3D point cloud data, the feature extractor is exposed to a wider range of spatial variations in training, which not only promotes the learning of more versatile and discriminative feature representation but also enhances the model's robustness in downstream tasks.

\subsection{Convolutional Transform Feature Network for Anomaly Detection}
\label{ppnet}
By utilizing the proxy classification task, the feature extractor has a versatile representation ability for  point clouds. The ability to capture multi-scale structural information in a lightweight manner is crucial for achieving high detection performance. To address this problem, we propose a lightweight CTF-Net for extracting multi-scale features in 3D anomaly detection. As shown in Figure~\ref{pipline}, it consists of several compositional convolution blocks (CCB) and Multilayer Perceptron (MLP). Specifically, each CCB performs the 1D convolutions with the kernel sizes of $3*1$ and $5*1$ on the input, respectively, and then concatenates the two convolution results to capture local information of each point at different scales. Besides, the 1D convolution with the kernel size of $1*1$ is applied to the concatenated features, followed by the Exponential Linear Units (ELU) activation function and batch normalization (BN). 

For any point cloud sample, CTF-Net uses four CCBs to enhance the dimensionality of each point to 512. Then the max operation is imposed on the feature map in the dimensionality of the sample size, and each point cloud sample is represented into a vector with a size of 512. Formally, this process can be described as:
\begin{equation}
\textbf{V}_{(1, 512)} = \max\{\text{CCB}_4 (\text{CCB}_{3} (\text{CCB}_{2} (\text{CCB}_1 (\textbf{P}_{(n, 3)})))\}.
\end{equation}

After that, the feature vector $\textbf{V}$ is mapped to a transformation matrix $\textbf{M}$ with a size of 3*3 by MLP, and the original point cloud \textbf{P}$_{(n, 3)}$ is multiplied with this transformation matrix \textbf{P}$_{(3, 3)}$ to obtain a better representation \textbf{F}$_{(n, 3)}$ of all points, which can be formally expressed as:
\begin{equation}
\textbf{F}_{(n, 3)} = \textbf{P}_{(n, 3)} \otimes MLP(\textbf{V}_{(1, 512)}).
\end{equation}

Subsequently, each transformed point is expanded to the dimensionality of 1024 by using multiple CCB, and each point cloud sample is mapped to a final feature vector with the dimensionality of 1024, which can be formulated as:
\begin{equation}
\textbf{F}_{(1, 1024)}^{*} = \max\{\text{CCB}_5 (\cdots(\text{CCB}_1 (\textbf{F}_{(n, 3)})))\}.
\end{equation}

With the carefully designed feature extraction network CTF-Net, we can train a model for 3D anomaly detection. During the training phase, the input point cloud data is first performed coordinate mapping by using the PCM module, and all points are divided into several groups through farthest point sampling. This process can be described as:
\begin{align}
    \textbf{p}_i &= FPS( PCM(\textbf{P}_{(n, 3 )})), \\
    \textbf{G}_i &= KNN(\textbf{p}_i),
\end{align}
where $\textbf{p}_i$ denotes the selected point by Farthest Point Sampling ($FPS$) from point clouds after PCM coordinate mapping, $\textbf{G}_i$ means the $i$-th group through searching $k$-nearest neighbors ($KNN$) for the point $\textbf{p}_i$.

For each group with $k$ points, the CTF-Net extracts features separately and the resulting features are stored in a memory bank:
\begin{equation}
   \textbf{Mem}_i = CTF\text{-}Net( \textbf{G}_i).
\end{equation}
Through this feature extraction process, we can learn the distribution of group features for normal samples to construct a memory bank for anomaly detection.

During the testing phase, the testing sample is similarly divided into several groups as in the training phase, and feature extraction is performed on each group separately. For each extracted group feature, its distance to the nearest neighbor in the memory bank is considered as the score, which can be calculated as:
\begin{equation}
    \textbf{S}_i= \parallel \textbf{Mem}_{near} - \textbf{F}_i \parallel,
\end{equation}
where $\textbf{F}_i$ and $\textbf{Mem}_{near}$ denote the extracted group feature and its nearest neighbor feature in the memory bank, respectively, and the symbol $\parallel \cdot \parallel$ represents the 2-norm. 
This score can be used as the pixel-level score to reflect the degree of abnormality of the selected point $\textbf{p}_i$ by FPS. While the maximum value of all point scores can be used to define the object-level score of the testing sample.

\begin{algorithm}[!ht]
\caption{Point Cloud Mapping (PCM)}
\label{PCMp}
\small
\begin{algorithmic}[1]
\Require Point cloud $\mathbf{P} = \{\mathbf{p}_i \in \mathbb{R}^3\}_{i=1}^n$
\Ensure Rotation-invariant representation $\mathbf{P}^* = \{\mathbf{p}_i^* \in \mathbb{R}^3\}_{i=1}^n$
\State Compute centroid: $\mathbf{c} \gets \frac{1}{n}\sum_{i=1}^n \mathbf{p}_i$ \hfill $\triangleright$ \parbox[t]{0.35\linewidth}{\raggedright Step 1: Center the point cloud}
\State Center points: $\widetilde{\mathbf{p}}_i \gets \mathbf{p}_i - \mathbf{c},\ \forall i \in \{1,\ldots,n\}$
\State Sort indices $I \gets \text{argsort}(\|\widetilde{\mathbf{p}}_i\|_2,\ \text{descending})$ \hfill $\triangleright$ \parbox[t]{0.35\linewidth}{\raggedright Step 2: Sort points by distance}
\State $\mathbf{u}_1 \gets \widetilde{\mathbf{p}}_{I[1]}$ \hfill $\triangleright$ \parbox[t]{0.35\linewidth}{\raggedright First basis vector candidate}
\For{$k \gets 2$ to $n$} \hfill $\triangleright$ \parbox[t]{0.35\linewidth}{\raggedright Step 3: Find linearly independent second vector}
    \If{$\text{rank}\left([\mathbf{u}_1, \widetilde{\mathbf{p}}_{I[k]}]\right) = 2$}
        \State $i_2 \gets I[k]$, $\mathbf{u}_2 \gets \widetilde{\mathbf{p}}_{i_2}$
        \State \textbf{break}
    \EndIf
\EndFor
\For{$k \gets n$ down to $1$} \hfill $\triangleright$ \parbox[t]{0.35\linewidth}{\raggedright Step 4: Find linearly independent third vector}
    \State $\mathbf{U} \gets [\mathbf{u}_1, \mathbf{u}_2, \widetilde{\mathbf{p}}_{I[k]}]$
    \If{$\text{rank}(\mathbf{U}) = 3$}
        \State $i_3 \gets I[k]$, $\mathbf{u}_3 \gets \widetilde{\mathbf{p}}_{i_3}$
        \State \textbf{break}
    \EndIf
\EndFor
\State Compute basis vector $\mathbf{e}_1 \gets \frac{\mathbf{u}_1}{\|\mathbf{u}_1\|_2}$ \hfill $\triangleright$ \parbox[t]{0.35\linewidth}{\raggedright Step 5: Gram-Schmidt orthogonalization}
\State $\mathbf{v}_2 \gets \mathbf{u}_2 - (\mathbf{u}_2^\top\mathbf{e}_1)\mathbf{e}_1$
\State Compute basis vector $\mathbf{e}_2 \gets \frac{\mathbf{v}_2}{\|\mathbf{v}_2\|_2}$
\State $\mathbf{v}_3 \gets \mathbf{u}_3 - \sum_{j=1}^2(\mathbf{u}_3^\top\mathbf{e}_j)\mathbf{e}_j$
\State Compute basis vector $\mathbf{e}_3 \gets \frac{\mathbf{v}_3}{\|\mathbf{v}_3\|_2}$
\State Compute transformation matrix $\mathbf{S} \gets [\mathbf{e}_1, \mathbf{e}_2, \mathbf{e}_3] \in \mathbb{R}^{3\times3}$ \hfill $\triangleright$ \parbox[t]{0.35\linewidth}{\raggedright Step 6: Apply transformation}
\State Transform points: $\mathbf{p}_i^* \gets \mathbf{S}^\top(\mathbf{p}_i - \mathbf{c}),\ \forall i \in \{1,\ldots,n\}$
\State \Return $\mathbf{P}^* = \{\mathbf{p}_i^*\}_{i=1}^n$
\end{algorithmic}
\end{algorithm}

\section{Experiments}
\label{experiment}
In this section, comparative experiments are first conducted to show the effectiveness of our method, followed by ablation studies and parameter sensitivity analysis to examine the rationality of the key components. Finally, extensibility experiments are also performed to demonstrate the generalization of our framework RIF with other feature extractors.

\subsection{Experimental Setup}
\textbf{Datasets.} We conducted comparative experiments on both available real or synthetic 3D AD datasets, namely Anomaly-ShapeNet and Real3D-AD.  {Anomaly-ShapeNet} has over 1,600 samples from 40 categories. The training set contains only four normal samples for each category, while the test set includes both normal and anomalous samples with diverse defects.  {Real3D-AD} is a high-resolution 3D dataset, containing 1,254 samples from 12 categories. The training set for each category includes four normal samples, while the test set for each category contains both normal and anomalous samples with different defects.

\noindent \textbf{Baselines.} Our method was compared with eight recently SOTA methods, including BTF~\cite{horwitz2022featureclassical3dfeatures},  M3DM~\cite{Wang2023multimodal}, PatchCore~\cite{roth2022totalrecallindustrialanomaly}, CPMF~\cite{cao2023complementarypseudomultimodalfeature}, R3D-AD~\cite{zhou2025r3dad}, Group3AD~\cite{zhu2024towards}, Reg3D-AD~\cite{Liu2023real3d}, MC3D-AD~\cite{cheng2025mc3dadunifiedgeometryawarereconstruction} and IMRNet~\cite{li2023scalable3danomalydetection}. The results of these methods are excerpted from their papers or implemented by publicly available codes.

\noindent \textbf{Evaluation Metrics.}
We used P-AUROC (Area Under the Receiver Operator Curve$\uparrow$) to evaluate pixel-level anomaly localisation accuracy and O-AUROC($\uparrow$) to assess object-level AD capability. Additionally, we adopted the metrics of P-AUPRO (Area Under the Per-Region-Overlap$\uparrow$) and O-AUPRO($\uparrow$) to evaluate pixel-level and object-level performance, respectively. 
Moreover, we employed Frames Per Second (FPS$\uparrow$) and Memory ($\downarrow$) to evaluate the inference speed and resource consumption. 

\noindent \textbf{Implementation Details.}
The proposed feature extractor CTF-Net was pre-trained from scratch on the ModelNet40 dataset, with each sample downsampling from 1024 to 512 points to match downstream tasks. During training and testing, the group size was set to 512 for feature extraction, with 512 points in each group. All codes were implemented in PyTorch using Python 3.8, and the experiments were conducted on a server equipped with an RTX 3090 (24GB).

\subsection{Experimental Results and Analysis}
To show the validity of our method, we conducted comparison experiments on Anomaly-ShapeNet and Real3D-AD, and the performance in terms of P-AUROC is shown in Table~\ref{AnomalyI} and Table~\ref{acf}, respectively. Prior work has demonstrated that rotation-invariant representation methods often exhibit diminished performance on general downstream tasks. In contrast, our proposed approach successfully reconciles this trade-off by achieving state-of-the-art AD performance while preserving robust rotation-invariant properties.

\begin{table*}[!htb]
  \centering
\resizebox{\textwidth}{!}{
    \begin{tabular}{l|cccccccccccccc}
    \toprule
    Method & cap0 & cap3 & helmet3 & cup0 & bowl4 & vase3 & headset1 & eraser0 & vase8 & cap4 & vase2 & vase4 & helmet0 & bucket1 \\
    \midrule
    BTF(Raw) & 0.524 & 0.687 & 0.700 & 0.632 & 0.563 & 0.602 & 0.475 & 0.637 & 0.550 & 0.469 & 0.403 & 0.613 & \multicolumn{1}{c}{0.504} & 0.686 \\
    BTF(FPFH) & {0.730} & 0.658 & \underline{0.724} & \underline{0.790} & 0.679 & {0.699} & 0.591 & 0.719 & 0.662 & 0.524 & 0.646 & 0.710 & 0.575 & 0.633 \\
    M3DM & 0.531 & 0.605 & 0.655 & 0.715 & 0.624 & 0.658 & 0.585 & 0.710 & 0.551 & 0.718 & \underline{0.737} & 0.655 & 0.599 & 0.699 \\
    PatchCore(FPFH) & 0.472 & 0.653 & \textbf{0.737} & 0.655 & 0.720 & 0.430 & 0.464 & 0.810 & 0.575 & 0.595 & 0.721 & 0.505 & 0.548 & 0.571 \\
    PatchCore(PointMAE) & 0.544 & 0.488 & 0.615 & 0.510 & 0.501 & 0.465 & 0.423 & 0.378 & 0.364 & 0.725 & 0.742 & 0.523 & 0.580 & 0.574 \\
    RegAD & 0.632 & {0.718} & 0.620 & 0.685 & {0.800} & 0.511 & 0.626 & {0.755} & {0.811} & {0.815} & 0.405 & {0.755} & 0.600 & 0.752 \\
    IMRNet & 0.715 & 0.706 & 0.663 & 0.643 & 0.576 & 0.401 & 0.476 & 0.548 & 0.635 & 0.753 & 0.614 & 0.524 & 0.598 & 0.774 \\
    CPMF & 0.601 & 0.551 & 0.520 & 0.497 & 0.683 & 0.582 & 0.458 & 0.689 & 0.529 & 0.553 & 0.582 & 0.514 & 0.555 & 0.601 \\
    R3D-AD & 0.666 & 0.676 & 0.708 & 0.600 & 0.651 & 0.595 & {0.676} & 0.660 & 0.597 & 0.678 & 0.712 & 0.674 & {0.623} & 0.615 \\
    {MC3D-AD} & \underline{{0.854}} & \underline{{0.903}} & {0.585} & {0.763} & {0.670} & \underline{{0.800}} & {0.592} & \underline{{0.820}} & \underline{{0.874}} & \underline{{0.858}} & \underline{{0.781}} & \underline{{0.772}} & \underline{{0.749}} & \textbf{{0.868}} \\
    {DUS-Net} & {0.701} & {0.763} & {0.682} & {0.727} & \textbf{{0.812}} & {0.731} & \underline{{0.749}} & {0.569} & {0.762} & {0.783} & {0.650} & {0.765} & {0.718} & {0.754} \\
    \rowcolor{gray!10} Ours & \textbf{0.968} & \textbf{0.955} & 0.719 & \textbf{0.912} & \underline{{0.802}} & \textbf{0.863} & \textbf{0.840} & \textbf{0.976} & \textbf{0.938} & \textbf{0.956} & \textbf{0.930} & \textbf{0.842} & \textbf{0.838} & \underline{{0.812}} \\
    \midrule
    \midrule
    Method & bottle3 & vase0 & bottle0 & tap1 & bowl0 & bucket0 & vase5 & vase1 & vase9 & ashtray0 & bottle1 & \multicolumn{1}{c}{tap0} & phone & cup1 \\
    \midrule
    BTF(Raw) & 0.720 & 0.618 & 0.551 & 0.564 & 0.524 & 0.617 & 0.585 & 0.549 & 0.564 & 0.512 & 0.491 & \multicolumn{1}{c}{0.527} & 0.583 & 0.561 \\
    BTF(FPFH) & 0.622 & 0.642 & 0.641 & 0.596 & 0.710 & 0.401 & 0.429 & 0.619 & 0.568 & 0.624 & 0.549 & \multicolumn{1}{c}{0.568} & 0.675 & 0.619 \\
    M3DM & 0.532 & 0.608 & 0.663 & 0.712 & 0.658 & {0.698} & 0.642 & 0.602 & 0.663 & 0.577 & 0.637 & \multicolumn{1}{c}{0.654} & 0.358 & 0.556 \\
    PatchCore(FPFH) & 0.512 & 0.655 & 0.654 & \textbf{{0.768}} & 0.524 & 0.459 & 0.447 & 0.453 & 0.663 & 0.597 & 0.687 & \multicolumn{1}{c}{\underline{0.733}} & 0.488 & 0.596 \\
    PatchCore(PointMAE) & 0.653 & 0.677 & 0.553 & 0.541 & 0.527 & 0.586 & 0.572 & 0.551 & 0.423 & 0.495 & 0.606 & \multicolumn{1}{c}{\textbf{0.858}} & {0.886} & \textbf{0.856} \\
    RegAD & 0.525 & 0.548 & {0.886} & {0.741} & {0.775} & {0.619} & 0.624 & 0.602 & {0.694} & {0.698} & 0.696 & \multicolumn{1}{c}{0.589} & 0.599 & 0.698 \\
    IMRNet & 0.641 & 0.535 & 0.556 & 0.699 & \underline{{0.781}} & 0.585 & {0.682} & \underline{0.685} & 0.691 & 0.671 & {0.702} & \multicolumn{1}{c}{0.681} & 0.742 & 0.688 \\
    CPMF & 0.435 & 0.458 & 0.521 & 0.657 & 0.745 & 0.486 & 0.651 & 0.486 & 0.545 & 0.615 & 0.571 & \multicolumn{1}{c}{0.458} & 0.545 & 0.509 \\
    R3D-AD & {0.750} & {0.743} & 0.722 & {0.716} & 0.711 & 0.593 & 0.642 & 0.616 & 0.617 & 0.689 & 0.674 & \multicolumn{1}{c}{0.643} & 0.693 & 0.617 \\
    {MC3D-AD} & \textbf{{0.902}} & \underline{{0.897}} & \underline{{0.902}} & {0.584} & {0.775} & \textbf{{0.902}} & {0.588} & {0.608} & \underline{{0.762}} & \underline{{0.807}} & \textbf{{0.867}} & \multicolumn{1}{c}{{0.502}} & \underline{{0.891}} & {0.694} \\
    {DUS-Net} & {0.641} & {0.699} & {0.749} & \underline{{0.743}} & {0.769} & \underline{{0.738}} & \textbf{{0.711}} & {0.648} & {0.581} & {0.612} & \underline{{0.822}} & \multicolumn{1}{c}{{0.728}} & {0.648} & {0.654} \\
    \rowcolor{gray!10} Ours & \underline{{0.813}} & \textbf{0.905} & \textbf{0.952} & 0.667 & \textbf{0.979} & 0.553 & \underline{{0.696}} & \textbf{0.807} & \textbf{0.775} & \textbf{0.865} & {0.710} & \multicolumn{1}{c}{0.670} & \textbf{0.955} & \underline{{0.831}} \\
    \midrule
    \midrule
    Method & vase7 & helmet2 & cap5 & shelf0 & bowl5 & bowl3 & helmet1 & bowl1 & headset0 & bag0 & bowl2 & jar & \multicolumn{2}{|c}{Mean} \\
    \midrule
    BTF(Raw) & 0.578 & 0.605 & 0.373 & 0.464 & 0.517 & {0.685} & 0.449 & 0.464 & 0.578 & 0.430 & 0.426 & 0.423 & \multicolumn{2}{|c}{0.550} \\
    BTF(FPFH) & 0.540 & 0.643 & 0.586 & 0.619 & 0.699 & 0.590 & \textbf{0.749} & \textbf{0.768} & 0.620 & {0.746} & 0.518 & 0.427 & \multicolumn{2}{|c}{0.628} \\
    M3DM & 0.517 & 0.623 & 0.655 & 0.554 & 0.489 & 0.657 & 0.427 & 0.663 & 0.581 & 0.637 & \underline{0.694} & 0.541 & \multicolumn{2}{|c}{0.616} \\
    PatchCore(FPFH) & 0.693 & 0.455 & \underline{0.795} & 0.613 & 0.358 & 0.327 & 0.489 & 0.531 & 0.583 & 0.574 & 0.625 & 0.478 & \multicolumn{2}{|c}{0.580} \\
    PatchCore(PointMAE) & 0.651 & 0.651 & 0.545 & 0.543 & 0.562 & 0.581 & 0.562 & 0.524 & 0.575 & 0.674 & 0.515 & 0.487 & \multicolumn{2}{|c}{0.577} \\
    RegAD & \underline{0.881} & \underline{0.825} & 0.467 & 0.688 & 0.691 & 0.654 & 0.624 & 0.615 & 0.580 & 0.715 & 0.593 & 0.599 & \multicolumn{2}{|c}{{0.668}} \\
    IMRNet & 0.593 & 0.644 & 0.742 & 0.605 & {0.715} & 0.599 & 0.604 & 0.705 & {0.705} & 0.668 & 0.684 & {0.765} & \multicolumn{2}{|c}{0.650} \\
    CPMF & 0.504 & 0.515 & 0.551 & \textbf{0.783} & 0.684 & 0.641 & 0.542 & 0.488 & 0.699 & 0.655 & 0.635 & 0.611 & \multicolumn{2}{|c}{0.573} \\
    R3D-AD & 0.619 & 0.598 & 0.662 & 0.681 & 0.572 & 0.633 & \underline{0.651} & \underline{0.742} & 0.637 & 0.654 & 0.550 & 0.736 & \multicolumn{2}{|c}{0.657} \\
    {MC3D-AD} & {0.576} & {0.818} & \underline{{0.882}} & {0.625} & {0.562} & \underline{{0.779}} & {0.591} & {0.562} & {0.666} & \underline{{0.857}} & {0.597} & \underline{{0.847}} & \multicolumn{2}{|c}{\underline{{0.748}}} \\
    {DUS-Net} & {0.728} & {0.744} & {0.844} & {0.740} & \underline{{0.744}} & {0.574} & \underline{{0.737}} & {0.735} & \underline{{0.718}} & {0.628} & {0.617} & {0.771} & \multicolumn{2}{|c}{{0.712}} \\
    \rowcolor{gray!10} Ours & \textbf{0.913} & \textbf{0.905} & \textbf{0.943} & \underline{0.765} & \textbf{0.927} & \textbf{0.872} & 0.639 & 0.613 & \textbf{0.895} & \textbf{0.957} & \textbf{0.859} & \textbf{0.967} & \multicolumn{2}{|c}{\textbf{0.845}} \\
    \bottomrule
    \end{tabular}%
}
\caption{P-AUROC performance of different methods on Anomaly-ShapeNet across 40 categories.}
\label{AnomalyI}%
\end{table*}%

\begin{table*}[!ht]
  \centering
  \resizebox{\textwidth}{!}{
    \begin{tabular}{l|cccccccccccc|c}
    \toprule
    \textbf{Method} & \textbf{Airplane} & \textbf{Car} & \textbf{Candybar} & \textbf{Chicken} & \textbf{Diamond} & \textbf{Duck} & \textbf{Fish} & \textbf{Gemstone} & \textbf{Seahorse} & \textbf{Shell} & \textbf{Starfish} & \textbf{Toffees} & \textbf{Mean} \\
    \midrule
    BTF(Raw) & 0.564 & 0.647 & 0.735 & 0.608 & 0.563 & 0.601 & 0.514 & 0.597 & 0.520 & 0.489 & 0.392 & 0.623 & 0.571 \\
    BTF(FPFH) & \textbf{0.738} & 0.708 & 0.864 & 0.693 & 0.882 & \textbf{0.875} & 0.709 & \underline{0.891} & 0.512 & 0.571 & 0.501 & 0.815 & 0.730 \\
    M3DM(PointBERT) & 0.523 & 0.593 & 0.682 & \textbf{0.790} & 0.594 & 0.668 & 0.589 & 0.646 & 0.574 & 0.732 & 0.563 & 0.677 & 0.636 \\
    M3DM(PointMAE) & 0.530 & 0.607 & 0.683 & 0.735 & 0.618 & 0.678 & 0.600 & 0.654 & 0.561 & 0.748 & 0.555 & 0.679 & 0.637 \\
    PatchCore(FPFH) & 0.471 & 0.643 & 0.637 & 0.618 & 0.760 & 0.430 & 0.464 & 0.830 & 0.544 & 0.596 & 0.522 & 0.411 & 0.577 \\
    PatchCore(FPFH+Raw) & 0.556 & 0.740 & 0.749 & 0.558 & 0.854 & 0.658 & 0.781 & 0.539 & 0.808 & 0.753 & 0.613 & 0.549 & 0.680 \\
    PatchCore(PointMAE) & 0.579 & 0.610 & 0.635 & 0.683 & 0.776 & 0.439 & 0.714 & 0.514 & 0.660 & 0.725 & 0.641 & 0.727 & 0.642 \\
    Reg3D-AD & 0.631 & 0.718 & 0.724 & 0.676 & 0.835 & 0.503 & 0.826 & 0.545 & 0.817 & 0.811 & 0.617 & 0.759 & 0.705 \\
    R3D-AD & 0.594 & 0.557 & 0.593 & 0.620 & 0.555 & 0.635 & 0.573 & 0.668 & 0.562 & 0.578 & 0.608 & 0.568 & 0.592 \\
    CPMF & 0.618 & \textbf{0.836} & 0.734 & 0.559 & 0.753 & 0.719 & \textbf{0.988} & 0.449 & \textbf{0.962} & 0.725 & \textbf{0.800} & \textbf{0.959} & 0.758 \\
    {Group3AD} & {0.636} & {0.745} & {0.738} & {\underline{0.759}} & {0.862} & {0.631} & {0.836} & {0.564} & {\underline{0.827}} & {\textbf{0.798}} & {0.625} & {0.803} & {0.735} \\
    MC3D-AD & 0.628 & \underline{0.819} & \underline{0.910} & 0.640 & \underline{0.942} & 0.822 & \underline{0.932} & 0.458 & 0.659 & \underline{0.778} & \underline{0.690} & \underline{0.934} & \underline{0.768} \\
    \rowcolor{gray!10} Ours & \underline{0.733} & 0.777 & \textbf{0.913} & 0.728 & \textbf{0.975} & \underline{0.869} & 0.710 & \textbf{0.935} & 0.573 & 0.714 & 0.588 & 0.888 & \textbf{0.784} \\
    \bottomrule
    \end{tabular}%
    }
    \caption{P-AUROC performance of different methods on Real3D-AD across 12 categories, where the best and second-place results are \textbf{boldfaced} and \underline{underlined}, respectively.}
  \label{acf}%
\end{table*}



 {1) Comparison on Anomaly-ShapeNet. As shown in Table 2, the proposed method achieved the best average P-AUROC of 84.5\%, which is improved over the competitive approach Reg3D-AD by 17.7\%, DUS-Net by 13.3\%, and R3D-AD by 18.8\%. Notably, for the ``cap0'' and ``jar'' categories, the P-AUROC values attained by our method are 96.8\% and 96.7\%, respectively, whereas Reg3D-AD only achieved 73.0\% and 76.5\%.} This significant improvement could potentially be attributed to the ability of the proposed method to adapt samples with variations using a unified feature representation, thereby enhancing the discriminative power of the extracted features, particularly in sparse point clouds. Furthermore, the utilization of a large number of points for each group ($K = 512$) may have contributed to the superior anomaly detection performance in such sparse point cloud scenario.
 
 {2) Comparison on Real3D-AD.} Table~\ref{acf} shows the pixel-level detection performance of all methods on Real3D-AD. Our method achieved 78.3\% P-AUROC on average. Among all selected methods, ours performs the best in terms of P-AUROC, surpassing the second-place method by 1.6\%. Particularly, for the ``Candybar'' and ``Diamond'' categories, the P-AUROC values of our method are 91.3\% and 97.5\%, respectively. The reason for this may be attributed to different point clouds sharing similar structures may have large variations in orientation and position, leading to large differences in the extracted features, while our method exhibits great robustness in capturing such variations compared to other methods. Moreover, for large-scale point clouds like Real3D-AD, our method obtained superior representation and performance with fewer groups ($G=512$) compared to Reg3D-AD ($G=16384$) and IMRNet (downsampled to 8192 points). This suggests that our CTF-Net could extract better features for 3D anomaly detection.

\subsection{Ablation Studies and Parameter Analysis}
{
To evaluate the importance of key components and the effect of parameters on our method, we conducted ablation studies and parameter sensitivity analysis on Anomaly-ShapeNet, and the results are shown in Table~\ref{ablation}, where the performance, efficiency, and memory consumption, including both the percentage and the absolute values of memory usage, are reported.}

\begin{table}[th!]
  \centering
  \resizebox{0.95\columnwidth}{!}{%
\begin{tabular}{l|cccc|ccc}
\toprule
\textbf{Method} & \textbf{P-AUROC} & \textbf{O-AUROC} & \textbf{P-AUPRO} & \textbf{O-AUPRO} & \textbf{FPS} & \textbf{Memory(\%)} & {\textbf{Memory(GB)}} \\
\midrule
\textbf{Ours$_{w/o\ \text{PCM}}$} & 0.5858 & 0.5028 & 0.0167 & 0.5196 & 0.31 & 25.85\% & {6.20} \\
\textbf{Ours$_{w/o\ \text{S3DA}}$} & 0.7884 & 0.6221 & 0.1732 & 0.6982 & 0.31 & 25.85\% & {6.20} \\
\textbf{Ours$_{replace\ \text{RANSAC}}$} & 0.7674 & 0.6732 & 0.1660 & 0.7260 & 0.29 & 25.85\% & {6.20} \\
\textbf{Ours$_{replace\ \text{PCA}}$} & 0.6141 & 0.5148 & 0.0178 & 0.5374 & 0.14 & 25.85\% & {6.20} \\
\hline
\textbf{Ours$_{w/o\ \text{Trans}}$} & 0.7793 & 0.6720 & 0.1622 & 0.7184 & 0.41 & 19.47\% & {4.67} \\
\textbf{Ours$_{replace\ \text{CCB}}$} & 0.8014 & 0.6213 & 0.1874 & 0.6844 & 0.50 & 16.98\% & {4.08} \\
\textbf{Ours$_{CCB=1}$} & 0.7263 & 0.6210 & 0.1364 & 0.6824 & 0.42 & 19.11\% & {4.59} \\
\textbf{Ours$_{CCB=2}$} & 0.7588 & 0.6223 & 0.1410 & 0.6963 & 0.39 & 19.94\% & {4.79} \\
\textbf{Ours$_{CCB=3}$} & 0.8001 & 0.6327 & 0.1684 & 0.7163 & 0.36 & 21.92\% & {5.26} \\
\textbf{Ours$_{CCB=4}$} & 0.8253 & 0.6609 & 0.2029 & 0.7432 & 0.35 & 23.33\% & {5.60} \\
\textbf{Ours$_{CCB=5}$} & \textbf{0.8446} & \textbf{0.6893} & \textbf{0.2124} & \textbf{0.7480} & 0.31 & 25.85\% & {6.20} \\
\hline
\textbf{Ours$_{G=64, K=512}$} & 0.6694 & 0.5960 & 0.0553 & 0.6655 & 0.50 & 16.34\% & {3.92} \\
\textbf{Ours$_{G=128, K=512}$} & 0.7456 & 0.6070 & 0.1021 & 0.6826 & 0.43 & 19.68\% & {4.72} \\
\textbf{Ours$_{G=256, K=512}$} & 0.8117 & 0.6692 & 0.1639 & 0.7318 & 0.36 & 20.43\% & {4.90} \\
\textbf{Ours$_{G=512, K=64}$} & 0.7919 & 0.6127 & 0.1193 & 0.6788 & 1.54 & 12.23\% & {2.94} \\
\textbf{Ours$_{G=512, K=128}$} & 0.7782 & 0.6297 & 0.0973 & 0.6809 & 1.02 & 14.42\% & {3.46} \\
\textbf{Ours$_{G=512, K=256}$} & 0.8215 & 0.6487 & 0.1532 & 0.7087 & 0.68 & 16.84\% & {4.04} \\
\textbf{Ours$_{G=512, K=512}$} & \textbf{0.8446} & \textbf{0.6893} & \textbf{0.2124} & \textbf{0.7480} & 0.31 & 25.85\% & {6.20} \\
\bottomrule
\end{tabular}
}
  \caption{Results of ablation and parameter sensitivity experiments, where  {w/o PCM},  {w/o S3DA}, and  {w/o Trans}. mean the proposed method without PCM, S3DA, and transformation matrix, respectively,  {replace RANSAC},  {replace PCA}, and  {replace CCB} mean replacing our PCM with RANSAC or PCA and replacing CCB with 1D convolution, respectively; Parameter  {CCB} denotes the number of blocks in the feature extractor CTF-Net, $G$ and $K$ represent the number of groups and the number of points within each group, respectively. Frames Per Second (FPS) and Memory (\% {and GB}) stand for the average inference speed and memory consumption of our CTF-Net with an RTX3090, respectively.} 
  \label{ablation}
\end{table}

Our method without the PCM component obtained only 58.58\% P-AUROC, 50.28\% O-AUROC, 1.67\% P-AUPRO, and 51.96\% O-AUPRO. However, after introducing the PCM component, our method ($G=512, K=512$) achieved a performance improvement by 25.88\%, 18.65\%, 19.57\%, and 22.84\% in terms of different evaluation metrics, respectively. The replacement of the PCM with the RANSAC registration in Reg3D-AD and commonly used PCA registration results in a significant degradation in performance. For instance, the PCA registration caused a 23.05\% degradation in P-AUROC, 17.45\% in O-AUROC, 19.46\% in P-AUPRO, and 21.05\% in O-AUPRO, respectively. The reason for this may be attributed to the fact that the PCM maps point clouds to a rotationally invariant space before feature extraction. For other methods, slight changes of point clouds in direction cause variations in extracted features, while the PCM eliminates the effect of orientation on data, thereby resulting in rotation-invariant features. This significantly improves the robustness of our method for 3D anomaly detection.

The proposed feature extractor CTF-Net, containing several CCB blocks and transformation matrix, was trained with the S3DA strategy. After removing S3DA, the performance was reduced by 5.62\%, 6.72\%, 3.92\%, and 4.98\%, respectively, in four metrics. This may be due to the fact that S3DA provides diverse training data to facilitate CTF-Net to learn more discriminative features.
When removing the transformation matrix, the performance was decreased by 6.53\%, 1.73\%, 5.02\%, and 2.96\%, respectively, in different metrics. Moreover, after replacing the CCB module with 1D convolution, the performance was reduced by 4.32\%, 6.80\%, 2.50\%, and 6.36\%, respectively. 
Finally, reducing the number of CCB modules from 5 to 1 resulted in performance decrease of 11.83\%, 6.83\%, 7.60\% and 6.56\%, respectively, in four metrics. The reason may be that more CCB modules benefit the capture of multi-scale structural features.

Our method involves two adjustable parameters: the number of groups $G$ and the number of points $K$ within each group. When varying the value of $G$ from 64 to 512, the performance of our method increased greatly.  Its P-AUROC, O-AUROC, P-AUPRO, and O-AUPRO were improved by 17.52\%, 9.33\%, 15.71\%, and 8.25\%, respectively. The explanation for this is that, as the number of groups grows, the number of features used to characterise each sample also increases, allowing for a more detailed description of the point cloud structure. 
When varying the value of $K$ from 64 to 512, the P-AUROC, O-AUROC, P-AUPRO, and O-AUPRO of our method improved by 5.27\%, 7.66\%, 9.31\%, and 6.92\%, respectively. This performance improvement can be attributed to the fact that, as the number of points in each group increases, the local information hidden within the points is further exploited, enabling a more comprehensive understanding of the regional structure and its surrounding context within the point cloud. While the setting of $G=512$ and $K=512$ achieved a balance between computational efficiency and detection precision.

{Our CTF-Net contains only 4.1M parameters, significantly fewer than existing alternatives such as PointMLP (12.6M), and requires only 6.2 GB of GPU memory on an RTX3090, which is approximately 0.9 GB less than memory-bank–based methods such as Reg3D-AD (7.1 GB). In addition, our approach achieved an inference speed of 0.33 FPS, outperforming methods like Reg3D-AD, which reports only 0.13 FPS.}

{\subsection{Robustness Experimental Results to Noisy Data.}
The complexity of environments and equipment instability may result in scanned point clouds containing noise in real-world scenarios. To evaluate the performance of our model with noise, we conducted comparative experiments on point clouds with different standard Gaussian noises under one-shot conditions. For the fairness comparison, all pre-training was performed on ModelNet40~\cite{sun2022modelnet40}. The results for Gaussian noise with a standard deviation of 0.002 and 0.003 are shown in Tables~\ref{os1} and \ref{os2}, respectively. It is observed that the performance decreases only slightly as the standard deviation of the noise increases, and is more noise-resistant compared to other approaches, showing better anomaly detection performance(P-AUROC at standard deviations of 0.02 and 0.03 was 9.8\% and 4.8\% higher than the baseline model Reg3D-AD~\cite{Liu2023real3d}, respectively). This may be attributed to the ability of rotationally invariant features to capture features more robustly under noisy conditions. These empirical results demonstrate the robustness of our method in detecting anomalies from noisy point clouds.}

\begin{table*}[!ht]
    \small
  \centering
\resizebox{\textwidth}{!}{
        \begin{tabular}{c|ccccccccccccc}
    \toprule
    \multicolumn{14}{c}{\textbf{(a) P-AUROC}} \\
    \midrule
    \textbf{Method} & \textbf{Airplane} & \textbf{Car} & \textbf{Candybar} & \textbf{Chicken} & \textbf{Diamond} & \textbf{Duck} & \textbf{Fish} & \textbf{Gemstone} & \textbf{Seahorse} & \textbf{Shell} & \textbf{Starfish} & \textbf{Toffees} & \textbf{Average} \\
    \midrule
    \textbf{M3DM(PointBERT)} & 0.512 & 0.574 & 0.624 & 0.734 & 0.582 & 0.642 & 0.557 & 0.643 & 0.532 & 0.648 & 0.443 & 0.657 & 0.596 \\
    \textbf{M3DM(PointMAE)} & 0.532 & 0.517 & 0.698 & 0.684 & 0.598 & 0.661 & 0.587 & 0.634 & 0.572 & 0.674 & 0.534 & 0.562 & 0.604 \\
    \textbf{PatchCore(PointMAE)} & 0.547 & 0.608 & 0.684 & 0.667 & 0.725 & 0.487 & 0.662 & 0.538 & 0.692 & \textbf{0.744} & 0.628 & 0.662 & 0.637 \\
    \textbf{Reg3D-AD} & 0.654 & 0.711 & 0.652 & 0.637 & 0.758 & 0.533 & \textbf{0.749} & 0.513 & \textbf{0.738} & 0.743 & \textbf{0.632} & 0.722 & 0.670 \\
    \textbf{RIF (CTF-Net) (Ours)} & \textbf{0.703} & \textbf{0.784} & \textbf{0.873} & \textbf{0.742} & \textbf{0.956} & \textbf{0.827} & 0.732 & \textbf{0.847} & 0.554 & 0.692 & 0.624 & \textbf{0.887} & \textbf{0.768} \\
    \midrule
    \multicolumn{1}{c}{} &       &       &       &       &       &       &       &       &       &       &       &       &  \\
    \midrule
    \multicolumn{14}{c}{\textbf{(b) O-AUROC}} \\
    \midrule
    \textbf{Method} & \textbf{Airplane} & \textbf{Car} & \textbf{Candybar} & \textbf{Chicken} & \textbf{Diamond} & \textbf{Duck} & \textbf{Fish} & \textbf{Gemstone} & \textbf{Seahorse} & \textbf{Shell} & \textbf{Starfish} & \textbf{Toffees} & \textbf{Average} \\
    \midrule
    \textbf{M3DM(PointBERT)} & 0.387 & 0.422 & 0.495 & 0.575 & 0.427 & 0.434 & 0.487 & 0.557 & 0.502 & 0.537 & 0.508 & 0.570 & 0.492 \\
    \textbf{M3DM(PointMAE)} & 0.527 & \textbf{0.508} & 0.482 & 0.560 & 0.612 & 0.438 & 0.530 & \textbf{0.624} & 0.512 & \textbf{0.688} & 0.520 & 0.562 & 0.547 \\
    \textbf{PatchCore(PointMAE)} & \textbf{0.711} & 0.501 & 0.547 & \textbf{0.797} & 0.757 & 0.514 & 0.578 & 0.411 & 0.504 & 0.528 & 0.524 & 0.560 & 0.578 \\
    \textbf{Reg3D-AD} & 0.602 & \textbf{0.638} & 0.604 & 0.668 & 0.768 & 0.490 & \textbf{0.677} & 0.344 & \textbf{0.610} & 0.538 & 0.469 & 0.588 & \textbf{0.583} \\
    \textbf{RIF (CTF-Net) (Ours)} & 0.608 & 0.500 & \textbf{0.624} & 0.566 & \textbf{0.824} & \textbf{0.550} & 0.524 & 0.557 & 0.501 & 0.492 & \textbf{0.580} & \textbf{0.666} & \textbf{0.583} \\
    \bottomrule
    \end{tabular}%
}
\caption{
One-Shot performance of different methods on Real3D-AD across 12 categories under Gaussian noise with a standard deviation of 0.002, where the best results are \textbf{boldfaced}.
}
  \label{os1}%
\end{table*}%

\begin{table*}[!ht]
\small
  \centering
\resizebox{\textwidth}{!}{
        \begin{tabular}{c|ccccccccccccc}
    \toprule
    \multicolumn{14}{c}{\textbf{(a) P-AUROC}} \\
    \midrule
    \textbf{Method} & \textbf{Airplane} & \textbf{Car} & \textbf{Candybar} & \textbf{Chicken} & \textbf{Diamond} & \textbf{Duck} & \textbf{Fish} & \textbf{Gemstone} & \textbf{Seahorse} & \textbf{Shell} & \textbf{Starfish} & \textbf{Toffees} & \textbf{Average} \\
    \midrule
    \textbf{M3DM(PointBERT)} & 0.501 & 0.511 & 0.594 & 0.701 & 0.574 & 0.622 & 0.501 & 0.597 & 0.501 & 0.587 & 0.401 & 0.524 & 0.551 \\
    \textbf{M3DM(PointMAE)} & 0.524 & 0.527 & 0.641 & 0.617 & 0.544 & 0.641 & 0.524 & 0.599 & 0.562 & 0.670 & 0.530 & 0.559 & 0.578 \\
    \textbf{PatchCore(PointMAE)} & 0.533 & 0.574 & 0.633 & 0.614 & 0.711 & 0.492 & 0.627 & 0.527 & 0.677 & \textbf{0.717} & \textbf{0.602} & 0.622 & 0.611 \\
    \textbf{Reg3D-AD} & \textbf{0.658} & 0.691 & 0.668 & 0.640 & 0.718 & 0.528 & 0.697 & 0.527 & \textbf{0.698} & 0.699 & 0.597 & 0.723 & 0.654 \\
    \textbf{RIF (CTF-Net) (Ours)} & 0.644 & \textbf{0.746} & \textbf{0.804} & \textbf{0.718} & \textbf{0.847} & \textbf{0.762} & \textbf{0.708} & \textbf{0.714} & 0.478 & 0.617 & 0.600 & \textbf{0.789} & \textbf{0.702} \\
    \midrule
        \multicolumn{1}{c}{} &       &       &       &       &       &       &       &       &       &       &       &       &  \\
        \midrule
    \multicolumn{14}{c}{\textbf{(b) O-AUROC}} \\
    \midrule
    \textbf{Method} & \textbf{Airplane} & \textbf{Car} & \textbf{Candybar} & \textbf{Chicken} & \textbf{Diamond} & \textbf{Duck} & \textbf{Fish} & \textbf{Gemstone} & \textbf{Seahorse} & \textbf{Shell} & \textbf{Starfish} & \textbf{Toffees} & \textbf{Average} \\
    \midrule
    \textbf{M3DM(PointBERT)} & 0.400 & 0.418 & 0.463 & 0.548 & 0.458 & 0.455 & 0.472 & 0.548 & 0.479 & 0.501 & 0.501 & 0.562 & 0.484 \\
    \textbf{M3DM(PointMAE)} & 0.501 & 0.504 & 0.472 & 0.547 & 0.601 & 0.452 & 0.517 & \textbf{0.576} & 0.520 & \textbf{0.648} & 0.498 & 0.542 & 0.532 \\
    \textbf{PatchCore(PointMAE)} & \textbf{0.681} & 0.508 & 0.532 & \textbf{0.728} & 0.648 & 0.524 & 0.568 & 0.428 & 0.502 & 0.542 & 0.496 & 0.543 & 0.558 \\
    \textbf{Reg3D-AD} & 0.578 & \textbf{0.640} & 0.587 & 0.643 & 0.614 & 0.501 & \textbf{0.657} & 0.341 & \textbf{0.594} & 0.501 & 0.428 & 0.571 & 0.555 \\
    \textbf{RIF (CTF-Net) (Ours)} & 0.617 & 0.517 & \textbf{0.630} & 0.527 & \textbf{0.727} & \textbf{0.552} & 0.530 & 0.560 & 0.508 & 0.504 & \textbf{0.542} & \textbf{0.618} & \textbf{0.569} \\
    \bottomrule
    \end{tabular}%
}
\caption{One-Shot performance of different methods on Real3D-AD across 12 categories under Gaussian noise with a standard deviation of 0.003, where the best results are \textbf{boldfaced}.}
  \label{os2}%
\end{table*}%

\subsection{Generalization of Our Framework RIF}
To examine the RIF's generalization, we conducted extensibility experiments on Anomaly-ShapeNet. The feature extractor CTF-Net was replaced by other representative point cloud feature extractors, including PointNet~\cite{Qi_2017_CVPR}, PointMLP~\cite{pointmae}, DGCNN~\cite{dgcnn}, PointMAE~\cite{pointmae}, Stratified-Transformer~\cite{stratifiedtransformer}, and PointTransformer-v3~\cite{pointtransformerv3}, and some results are reported in Table~\ref{generalization1}.
{For transformer-based methods, which are capable of processing all patches in parallel, we adopt the evaluation protocol of MC3D-AD and utilize 10{,}000 patches to ensure a fair comparison. In contrast, convolution-based baselines and our proposed approach process patches in a sequential manner due to their architectural characteristics.}

\begin{table}[!ht]
\small
  \centering
  \resizebox{0.8\columnwidth}{!}{
    \begin{tabular}{c|cccc}
    \toprule
    \textbf{Method} & \textbf{P-AUROC} & \textbf{O-AUROC} & \textbf{P-AUPRO} & \textbf{O-AUPRO} \\
    \midrule
    \textbf{RIF$_{\text{w/o PCM}}$ + PointNet} & 0.5081  & 0.5037  & 0.0177  & 0.5716  \\
    \textbf{RIF$_{\text{PCM}}$ + PointNet} & \textbf{0.8348}  & 0.6770  & 0.1964  & 0.7336  \\
    \textbf{RIF$_{\text{w/o PCM}}$ + PointMLP} & 0.7433  & 0.5751  & 0.0679  & 0.6418  \\
    \textbf{RIF$_{\text{PCM}}$ + PointMLP} & 0.8196  & \textbf{0.7533}  & \textbf{0.2987}  & 0.7950  \\
    \textbf{RIF$_{\text{w/o PCM}}$ + DGCNN} & 0.6054  & 0.5200  & 0.0230  & 0.5857  \\
    \textbf{RIF$_{\text{PCM}}$ + DGCNN} & 0.7658  & 0.7170  & 0.1963  & \textbf{0.8389}  \\
    \textbf{RIF$_{\text{w/o PCM}}$ + PointMAE} &0.6442 &0.5897&0.0437&0.6148\\
    \textbf{RIF$_{\text{PCM}}$ + PointMAE}&0.7242&0.6880&0.0544&0.7522 \\
    \textbf{RIF$_{\text{w/o PCM}}$ + Stratified-Transformer}&0.6142&0.6142&0.0208&0.6428 \\
    \textbf{RIF$_{\text{PCM}}$ + Stratified-Transformer}&0.6357&0.6685&0.0574&0.7125 \\
    \textbf{RIF$_{\text{w/o PCM}}$ + PointTransformer-v3}&0.6328&0.5927&0.0411&0.6284 \\
    \textbf{RIF$_{\text{PCM}}$ + PointTransformer-v3}&0.7842&0.7110&0.2384&0.7842 \\
    \bottomrule
    \end{tabular}%
  }
  \caption{Results of RIF with other feature extractors on Anomaly-ShapeNet in terms of different evaluation metrics.}
  \label{generalization1}
\end{table}

As shown in Table \ref{generalization1}, the integration of RIF with other feature extractors greatly enhances anomaly detection performance. On Anomaly-ShapeNet, RIF with PCM achieved notable improvements over baseline (w/o PCM): PointNet (P-AUROC: +32.67\%, O-AUROC: +17.33\%), PointMLP (P-AUROC: +7.63\%, O-AUROC: +17.82\%), DGCNN (P-AUROC: +16.04\%, O-AUROC: +19.70\%), PointMAE (P-AUROC: +8.00\%, O-AUROC: +9.83\%), Stratified-Transformer (P-AUROC: +2.15\%, O -AUROC: +5.43\%), and PointTransformer-v3 (P-AUROC: +15.14\%, O-AUROC: +11.83\%). 

For Real3D-AD, as shown in Table \ref{generalization2}, RIF with PCM obtained significant performance improvement, particularly with PointNet (P-AUROC: +24.23\%, O-AUROC: +8.80\%), DGCNN (P-AUROC: +16.17\%, O -AUROC: +3.15\%), Stratified-Transformer (P-AUROC: +10.23\%, O -AUROC: +3.14\%), and Point Transformer-v3 (P-AUROC: +6.12\%, O-AUROC: +2.30). These results demonstrate the strong adaptability of RIF to convolution-based (PointNet, DGCNN), MLP-based (PointMLP), and transformer-based (PointMAE, Stratified-Transformer, PointTransformer-v3) architectures, making it highly suitable for industrial 3D anomaly detection scenarios where available data exhibits a high degree of diversity and variability.

\begin{table}[!ht]
\small
  \centering
  \resizebox{0.8\columnwidth}{!}{
    \begin{tabular}{c|cccc}
    \toprule
    \multicolumn{5}{c}{\textbf{Real3D-AD}} \\
    \midrule
    \textbf{Method} & \textbf{P-AUROC} & \textbf{O-AUROC} & \textbf{P-AUPRO} & \textbf{O-AUPRO} \\
    \midrule
    \textbf{RIF$_{\text{w/o PCM}}$ + PointNet} & 0.5387  & 0.5132  & 0.0171  & 0.5433  \\
    \textbf{RIF$_{\text{PCM}}$ + PointNet} & \textbf{0.7810}  & 0.6012  & \textbf{0.1053}  & 0.6035  \\
    \textbf{RIF$_{\text{w/o PCM}}$ + PointMLP} & 0.7267  & 0.5365  & 0.0483  & 0.5693  \\
    \textbf{RIF$_{\text{PCM}}$ + PointMLP} & 0.7673  & 0.5194  & 0.0548  & 0.5458  \\
    \textbf{RIF$_{\text{w/o PCM}}$ + DGCNN} & 0.5962  & 0.4932  & 0.0273  & 0.5200  \\
    \textbf{RIF$_{\text{PCM}}$ + DGCNN} & 0.7579  & 0.5247  & 0.0552  & 0.5403  \\
    \textbf{RIF$_{\text{w/o PCM}}$ + PointMAE} &0.6214 &0.5430 &0.0355 &0.5398\\
    \textbf{RIF$_{\text{PCM}}$ + PointMAE}&0.6478 &0.5882 &0.0398 &0.6112 \\
    \textbf{RIF$_{\text{w/o PCM}}$ + Stratified-Transformer}&0.6447&0.6128&0.0543&0.6427 \\
    \textbf{RIF$_{\text{PCM}}$ + Stratified-Transformer}&0.7470&\textbf{0.6443}&0.0830&\textbf{0.6574} \\
    \textbf{RIF$_{\text{w/o PCM}}$ + PointTransformer-v3}&0.6148&0.5883&0.0361&0.6044 \\
    \textbf{RIF$_{\text{PCM}}$ + PointTransformer-v3}&0.6760&0.6113&0.0743&0.6434 \\
    \bottomrule
    \end{tabular}%
  }
  \caption{Results of RIF with other feature extractors in terms of different evaluation metrics.}
  \label{generalization2}
\end{table}

\subsection{Efficiency Comparison between Our Method and Existing Methods}
\begin{figure}[!ht]
    \centering
    \includegraphics[width=0.70\linewidth]{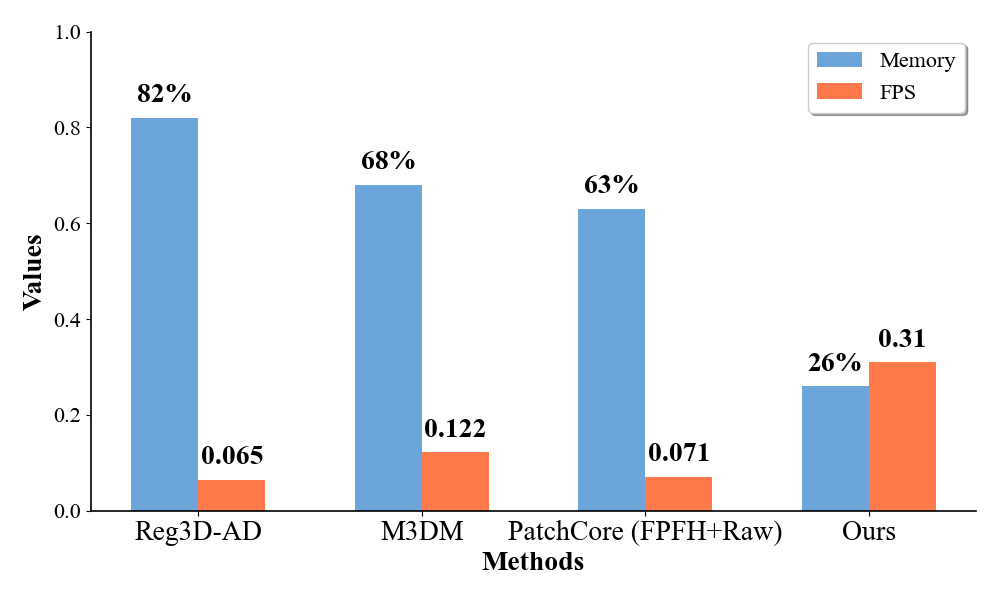}
    \caption{Comparison between our method with other methods in terms of inference speed and maximum memory during inference on Anomaly-ShapeNet, with experiments performed on an RTX3090.}
    \label{speed}
\end{figure}

Inference speed and memory usage are very important in industrial applications~\cite{pmlr-v202-chu23b}. As shown in Figure~\ref{speed}, our method with the rotationally invariant feature mechanism requires only a low memory capacity to achieve the best results. Compared to existing feature embedding approaches, our approach outperforms in both memory consumption (Memory: 26\%) and inference speed (FPS: 0.31). This may be attributed to the use of a lightweight feature extraction network and fewer groups to construct the memory bank, with a significant reduction in the search space.

\subsection{Potential Data Leakage Issues}
ModelNet is used as a pre-training method in some baselines. To avoid potential data leakage, we utilized S3DIS~\cite{Armeni2016CVPR} for pre-training, and the results are shown in Table~\ref{tab:performance_comparison}. Similar detection results from pre-training on a large indoor dataset validate that the proposed method does not suffer from data leakage.
\begin{table}[htbp]
\centering
\begin{tabular}{lcc}
\toprule
Method & RIF$_{ModelNet40}$ & RIF$_{S3DIS}$ \\
\midrule
O-AUROC$_{Real3D\text{-}AD}$ & 0.796 & \textbf{0.809} \\
P-AUROC$_{Real3D\text{-}AD}$ &\textbf{0.784} & 0.782 \\
O-AUROC$_{Anomaly\text{-}ShapeNet}$ & 0.689 & \textbf{0.701} \\
P-AUROC$_{Anomaly\text{-}ShapeNet}$ & \textbf{0.845} & 0.840 \\
\bottomrule
\end{tabular}
\caption{Performance comparison on pre-training with different datasets.}
\label{tab:performance_comparison}
\end{table}
\section{Conclusion}
\label{Conclusion}
This paper presented an innovative 3D anomaly detection~(AD) framework RIF, aiming at addressing challenges from samples with orientation and position variations in real industrial production scenarios. To eliminate the negative effects of data variations on feature extraction, we proposed the Point Coordinate Mapping (PCM) technique to transform point clouds from any orientation into a consistent point cloud representation. Moreover, to learn from normal point cloud samples, we developed a Convolutional Transform Feature Network (CTF-Net) to derive highly discriminative features for the memory bank. Additionally, we introduced a Spatial 3D Data Augmentation (S3DA) technique to aid the pre-training of our feature extraction networks. Experiments on the Anomaly-ShapeNet and Real3D-AD datasets show that our method achieved SOTA performance across different metrics. Our RIF framework is also highly adaptable to other feature extraction networks, providing a promising solution for industrial 3D AD.

{\textbf{Limitation:} Although the proposed approach has yielded promising results, its patch-based representation imposes inherent limitations on achieving fine-grained, point-level predictive accuracy. As part of our future work, we aim to extend rotation-invariant reconstruction paradigms to enable more precise, point-wise anomaly localization.}

\section{Acknowledgments}
This work was supported in part by the National Natural Science Foundation of China (Nos. 62476171 and 62206122), Guangdong Basic and Applied Basic Research Foundation (No. 2024A1515011367), Guangdong-Macao Science and Technology Innovation Joint Fundation (No. 2024A1515011367), National Undergraduate Training Program for Innovation and Entrepreneurship (No. S202510590118), Guangdong Provincial Key Laboratory (No. 2023B1212060076), Tencent ``Rhinoceros Birds” - Scientific Research Foundation for Young Teachers of Shenzhen University, and the Signal, Information, and Biological System Processing Laboratory at Shenzhen Audencia Financial Technology Institute, Shenzhen University.

\bibliographystyle{ieeetr}
\bibliography{refs}
\end{document}